\documentclass[11pt]{article}
\usepackage{coling2020}
\usepackage{fancyhdr}
\usepackage{times}
\usepackage{url}
\usepackage{latexsym}
\usepackage[table,xcdraw]{xcolor}
\usepackage{times}
\usepackage{latexsym}

\usepackage{multirow}
\usepackage{adjustbox}
\usepackage{booktabs}
\usepackage{amssymb}
\usepackage{tablefootnote}
\usepackage{times}
\usepackage{fancyhdr,graphicx,amsmath,amssymb}
\usepackage[ruled,vlined]{algorithm2e}
\include{pythonlisting}

\usepackage{microtype}

\usepackage{tikz-dependency}
\usepackage{subcaption} 
\usepackage{booktabs}
\usepackage{graphicx}
\usepackage{makecell}
\usepackage{multirow}
\usepackage{adjustbox}
\usepackage{dsfont}

\newcommand\rBS{\raisebox{0ex}{\resizebox{1.1ex}{1.3ex}{\reflectbox{\texttt{/}}}}}
\newcommand\lSL{\raisebox{0ex}{\resizebox{1.1ex}{1.3ex}{\texttt{/}}}}
\newcommand\lT{\texttt{<}}
\newcommand\rT{\texttt{>}}

\colingfinalcopy 

\title{Bracketing Encodings for 2-Planar Dependency Parsing}
  
 \author{Michalina Strzyz \qquad David Vilares \qquad Carlos G\'omez-Rodr\'iguez\\
  Universidade da Coru\~na, CITIC \\
  FASTPARSE Lab, LyS Research Group \\
  Departamento de Ciencias de la Computación y Tecnologías de la Información\\
  Campus de Elvi\~na, s/n, 15071 A Coru\~na, Spain\\
  {\tt \{michalina.strzyz,david.vilares,carlos.gomez\}@udc.es}}

\date{}

\begin{document}
\maketitle
\begin{abstract}
We present a bracketing-based encoding that can be used to represent any 2-planar dependency tree over a sentence of length $n$ as a sequence of $n$ labels, hence providing almost total coverage of crossing arcs in sequence labeling parsing. First, we show that existing bracketing encodings for parsing as labeling can only handle a very mild extension of projective trees. Second, we overcome this limitation by taking into account the well-known property of 2-planarity, which is present in the vast majority of dependency syntactic structures in treebanks, i.e., the arcs of a dependency tree can be split into two planes such that arcs in a given plane do not cross. We take advantage of this property to design a method that balances the brackets and that encodes the arcs belonging to each of those planes, allowing for almost unrestricted non-projectivity ($\sim 99.9\%$ coverage) in sequence labeling parsing. The experiments show that our linearizations improve over the accuracy of the original bracketing encoding in highly non-projective treebanks (on average by $0.4$ LAS), while achieving a similar speed. Also, they are especially suitable when PoS tags are not used as input parameters to the models.

\end{abstract}

\blfootnote{This work is licensed under a Creative Commons Attribution 4.0 International Licence. Licence details: \url{http://creativecommons.org/licenses/by/4.0/}.}

\section{Introduction}

In the last few years, approaches that cast syntactic parsing as the task of finding a sequence have gained traction for both dependency and constituency parsing. In sequence-to-sequence (seq2seq) parsing \cite{Vinyals2015,li-etal-2018-seq2seq}, parse trees are represented as arbitrary-length sequences, where the attention mechanism can be seen as an abstraction of the stack and the buffer in transition-based systems that decides what words are relevant to make a decision at a given time step. In sequence labeling parsing \cite{gomez-rodriguez-vilares-2018-constituent,strzyz-etal-2019-viable}, the tree for a sentence of length $n$ is represented as a sequence of $n$ labels, one per word, so the parsing process is word-synchronous \cite{tetratagging} and can be addressed by frameworks traditionally used for other natural language processing tasks, such as part-of-speech tagging or named-entity recognition. Current sequence labeling parsers combine competitive accuracy with high computational efficiency, while providing extra simplicity using off-the-shelf sequence labeling software without the need for ad-hoc parsing algorithms.

In the realm of dependency parsing, pioneering work dates back to \newcite{Spoustova}, who used a relative PoS-tag based encoding to represent trees as label sequences, but the resulting accuracy was not practical even for the standards of the time, probably due to the inability of pre-deep-learning architectures to successfully learn the representation. Using more modern architectures with the ability to contextualize words based on the sentence, and various tree encodings, \newcite{strzyz-etal-2019-viable} were the first to show that competitive accuracy could be reached. 

Subsequently, this accuracy has been improved further by techniques like
the use of multi-task learning to parse dependencies and constituents together \cite{strzyz-etal-2019-sequence} and of contextualized embeddings \cite{VilStrSogGom2020a}.

\begin{figure}[] 

\begin{subfigure}{\columnwidth}
\centering
\begin{subfigure}[b]{0.5\linewidth}

\begin{dependency}[hide label, theme=simple]
\begin{deptext}[column sep=1em]
  $w_{0}$ \&$w_{1}$ \& $w_{2}$ \& $w_{3}$ \& $w_{4}$ \& $w_{5}$ \& $w_{6}$ \&  \\
 \texttt{root}\& $\emptyset$\&\lSL\lSL\lSL\rT\&\lSL\rT\&\lSL\rT\&\rT\&\rT \\
\end{deptext}

\depedge{2}{3}{}
\depedge{2}{4}{}
\depedge{2}{6}{}
\depedge{3}{5}{}
\depedge{4}{7}{}
\depedge{1}{2}{}
\end{dependency}
\end{subfigure}
\caption{Projective encoding restricted to a single plane. Infeasible to reconstruct a non-projective sentence.}
\label{fig:bracketing}
\centering
\begin{subfigure}[b]{0.5\linewidth}
\begin{dependency}[hide label, theme=simple]
\begin{deptext}[column sep=1em]
  $w_{0}$ \&$w_{1}$ \& $w_{2}$ \& $w_{3}$ \& $w_{4}$ \& $w_{5}$ \& $w_{6}$ \& 
    \\
  \texttt{root}\& $\emptyset$\&\lSL\lSL\lSL\rT\&\textcolor{red}{\bf $\lSL^{\bf *}$}\rT\&\textcolor{red}{\bf $\rT^{*}$}\&\rT\&$\emptyset$ \\
\end{deptext}
\depedge{2}{3}{}
\depedge{2}{4}{}
\depedge{2}{6}{}
\depedge[edge style={red,densely dotted}]{3}{5}{}
\depedge{1}{2}{}

\end{dependency}
\end{subfigure}
\caption{Non-projective 2-planar encoding with second-plane-averse greedy plane assignment. The arc $w_3 \rightarrow w_6$ is not assigned a plane because it would cross arcs belonging to both planes, which is forbidden by the 2-planar constraint.}
\label{fig:2-greedy}

\centering
\begin{subfigure}[b]{0.5\linewidth}
\begin{dependency}[hide label, theme=simple]
\begin{deptext}[column sep=1em]
 $w_{0}$ \& $w_{1}$ \& $w_{2}$ \& $w_{3}$ \& $w_{4}$ \& $w_{5}$ \& $w_{6}$ \& 
    \\
  \texttt{root} \& $\emptyset$\&\textcolor{red}{\bf $\lSL^{*}$}\lSL\lSL\rT\&\textcolor{red}{\bf $\lSL^{*}$}\rT\&\lSL\textcolor{red}{\bf $\rT^{*}$}\&\textcolor{red}{\bf $\rT^{*}$}\&\rT \\
\end{deptext}
\depedge{2}{3}{}
\depedge{2}{4}{}
\depedge[edge style={red,densely dotted}]{2}{6}{}
\depedge[edge style={red,densely dotted}]{3}{5}{}
\depedge{4}{7}{}
\depedge{1}{2}{}

\end{dependency}
\end{subfigure}
\caption{Non-projective 2-planar encoding with plane assignment based on restriction propagation on the crossings graph.}
\label{fig:2-prop}

\end{subfigure}
\caption{Bracketing-based encodings with their plane assignment strategies for a non-projective sentence. The red, dotted lines refer to the arcs represented in the second plane, denoted by * in the encoding label.}
\label{fig:fin}
\end{figure}

While parsing as sequence labeling does not need specific parsing algorithms or data structures, as in graph-based or transition-based parsing, the responsibility of providing suitable parsing representations with reasonable coverage and learnability falls instead on the encoding used to represent trees as sequences of labels. \newcite{strzyz-etal-2019-viable} used four different encodings that obtained substantially different parsing accuracies in the experiments, with two encodings 
achieving competitive accuracy: the relative PoS tag (rel-PoS) encoding of \newcite{Spoustova} and a new encoding based on balanced brackets, inspired by \newcite{yli-jyra-gomez-rodriguez-2017-generic}. 
While the encoding of \newcite{Spoustova} achieved a good accuracy,
and it has full coverage of non-projective dependency trees, it 
requires PoS tags to encode the
dependency arcs. This can be seen as a weakness, not just because computing and feeding PoS tags increases the latency, but also because 
the traditional assumption that PoS tagging is needed for parsing is being increasingly called into question \cite{de-lhoneux-etal-2017-raw,smith-etal-2018-investigation,kitaev-klein-2018-constituency,anderson2020frailty}. Low-frequency PoS tags can cause sparsity in the encoding, and low-quality PoS tags could be a potential source of errors in low-resource languages.

For this reason, 
\newcite{lacroix-2019-dependency} proposed two alternative encodings with the same relative indexing philosophy, but without using PoS tags. However, these encodings require a composition of two sequence labeling processes instead of one.

On the other hand, the bracketing encoding inspired in \cite{yli-jyra-gomez-rodriguez-2017-generic} represents the trees independently of PoS tags or any other previous tagging step, but it has the limitation of being restricted to 
a very mild extension of projective trees.

\paragraph{Contribution.} In this paper, we extend the idea of the bracketing-based encoding to non-projective parsing by defining a variant that can encode all 2-planar dependency trees \cite{Yli2003}. 2-planar dependency trees have been shown to cover the vast majority of non-projective trees in attested sentences \cite{gomez-rodriguez-2016-squibs} and have been used in transition-based parsing  \cite{gomez-rodriguez-nivre-2013-divisible,fernandez-gonzalez-gomez-rodriguez-2018-dynamic-oracle}. We show that our encoding provides better parsing accuracy than the original bracketing-based encoding on highly non-projective UD treebanks; and than the rel-PoS encoding when assuming PoS tags are not fed as input parameters to the models. The source code is available at \url{https://github.com/mstrise/dep2label}.

\section{Preliminaries}

Given a sentence $w_1 \ldots w_n$, we associate the words with nodes ${0,1,\ldots,n}$, where $0$ is a dummy root node. Then, a dependency graph is an edge-labeled graph $(V,E)$ with $V=\{0,1,\ldots,n\}$ and $E$ a set of edges of the form $(h,d,l)$ where $h \in V$ is the head, $d \in V \setminus \{0\}$ is the dependent, and $l$ is the dependency label. The goal of a dependency parser is to find a dependency graph that is a tree (i.e. without cycles, and with no dependent having more than one head) rooted at node $0$.

\subsection{Bracketing encoding}

Dependency arcs are encoded through a sequence of bracket elements from a set $B=\{\lT,\rBS,\lSL,\rT\}$.
A balanced pair of brackets $(\lT,\rBS)$ in the labels of the words $w_i$ and $w_j$ represents a left arc from word $w_j$ to $w_{i-1}$. A balanced pair of brackets $(\lSL,\rT)$ in the labels of the words $w_i$ and $w_j$ represents a right arc from word $w_{i-1}$ to $w_j$.
one incoming arc and several outgoing arcs, resulting on labels composed of several such brackets, following the regular expression \verb@(<)?((\)*|(/)*)(>)?@.
As shown in Figure \ref{fig:bracketing}, the token $w_{2}$ is assigned a label $\lSL\lSL\lSL\rT$ that can be interpreted as: the previous token $w_{1}$ has three outgoing arcs to the right and one of them matches the left incoming arc of $w_{2}$ (\lSL\rT) meaning that $w_{1}$ is the head of $w_{2}$. The remaining two dependents will be given by the
matching $\rT$ in the labels of the following words.

Since each opening bracket is always matched to the closest same-direction closing bracket, this encoding is unable to handle crossing arcs in the same direction. An attempt of encoding such crossing arcs will result in decoding into non-crossing arcs.However, the encoding can handle crossing arcs in opposite directions, as long as left and right brackets are balanced independently (e.g. by using separate stacks for each kind of bracket).
The paper by \newcite{strzyz-etal-2019-viable} erroneously describes the encoding as only supporting projective trees. In fact, 
the implementation in that paper is supporting this mild extension of projectivity where crossing arcs in opposite directions are allowed.

\subsection{2-Planarity}

\label{sec:2planarity}

A dependency graph $(V,E)$ is said to be $k$-planar, for $k \ge 1$, if there is a partition of the edges into sets $E_1, \ldots, E_k$, called planes, in such a way that edges that are in the same plane do not cross. For $k=1$, this corresponds to the concept of a noncrossing dependency graph \cite{kuhlmann-jonsson-2015-parsing} or planar linear arrangement \cite{Chao92} (not to be confused with a planar graph). Under the assumption of trees rooted at the dummy root node $0$, $1$-planar trees are equivalent to the well-known projective trees. For $k \ge 2$, this means that the dependency graph (together with the linear order of the words) is a $k$-page book embedding of a graph (see \cite{pitler-etal-2013-finding}). Intuitively, a $k$-planar graph is one where each arc can be assigned one out of $k$ colors in such a way that arcs with the same color do not cross (see Figure \ref{fig:fin}).

$2$-planarity has been shown to be particularly relevant for parsing, as the overwhelming majority of syntactic structures in syntactic treebanks has been shown to be 2-planar \cite{gomez-rodriguez-nivre-2013-divisible,gomez-rodriguez-2016-squibs} and efficient transition-based parsers have been proposed for this set of structures \cite{gomez-rodriguez-nivre-2010-transition,fernandez-gonzalez-gomez-rodriguez-2018-dynamic-oracle}.

\section{2-Planar bracketing encodings}

\label{sec:2planarencoding}
In order to support the extended non-projective coverage provided by 2-planarity in the bracketing system, we balance a different set of brackets for each plane.

We introduce a set of ``star'' bracket elements denoting arcs belonging to the second plane, $B^{*}=\{\lT^{*},\rBS^{*},\lSL^{*},\rT^{*}\}$. A token $w_{i}$ can be assigned elements from both $B$ and $B^{*}$. 
Brackets only match when they are on the same plane, i.e., $(\lT,\rBS)$, $(\lSL,\rT)$ are matching pairs of brackets that encode arcs in the first plane, and $(\lT^{*},\rBS^{*})$, $(\lSL^{*},\rT^{*})$ are matching pairs of brackets that encode arcs in the second plane. The decoding process is implemented by operating on separate stacks for the first-plane brackets and the second-plane brackets.

\subsection{Plane assignment strategies}

According to the definition in Section~\ref{sec:2planarity}, a tree is 2-planar if its edges can be partitioned into two planes, $E_1$ and $E_2$, such that edges in the same plane do not cross. However, often this partition is not unique (for example, in the case of trees that are also $1$-planar, \emph{any} partition satisfies the condition). Thus, for the encoding in Section~\ref{sec:2planarencoding} to provide a single sequence of labels for each gold tree during training, we need to fix a \emph{plane assignment strategy}, i.e., a canonical way of assigning each arc to a plane to obtain such a partition. While the number of possible partitions is exponential in the size of the tree, desirable partitions should be easily learnable, i.e., follow predictable patterns. Given that the amount of crossing dependencies in treebanks is scarce \cite{FerGomEstPhysA2018}, it makes sense to look for partitions that do not make use of an extra plane when not needed, 
so that the parsing of sentences or fragments without crossing arcs does not become more difficult or need more output labels than in the basic bracketing encoding (as they will only use one plane and thus one set of brackets).
Following this general principle, we define the following plane assignment strategies:

\paragraph{Second-Plane-Averse Greedy Plane Assignment} Arcs in the gold tree are traversed in left-to-right order of their right endpoint, with shortest arcs first when they share a right endpoint (this is the order in which arcs will be decoded using a stack, see Section \ref{sec:decoding}). For each arc $a$, we assign the first plane if possible (i.e., if no arc crossing $a$ has already been assigned the first plane). Otherwise, we assign the second plane if possible, or no plane if the arc $a$ crosses arcs assigned to both planes. The process is formally described with pseudocode in Algorithm~\ref{greedyalgo}.

\begin{figure}[ht]
  \centering
  \begin{minipage}{.7\linewidth}
  
    \begin{algorithm}[H]
\SetAlgoLined
  \KwIn{A set of arcs $T$, and input length $n$} 
  \KwResult{Two sets (planes) of arcs $P_1,P_2$}
$P_1 \gets \varnothing$\;
$P_2 \gets \varnothing$\;
 \For{$x_r \gets 1$ \textbf{to} $n$}{
    \For{$x_l \gets x_r-1$ \textbf{downto} $0$}{
       \If{$\exists a \in T \mid a = (x_l,x_r,l) \vee a = (x_r,x_l,l)$}{
         nextArc $\gets a$\;
         $C \gets \{ b \in (P_1 \cup P_2) \mid b$ crosses $a \}$\;
         \uIf{$C \cap P_1 = \varnothing$}{
            $P_1 \gets P_1 \cup \{ $nextArc$ \}$\;
           }
         \uElseIf{$C \cap P_2 = \varnothing$}{
            $P_2 \gets P_2 \cup \{ $nextArc$ \}$\; 
           }
          \uElse{
            do nothing (failed to assign nextArc to a plane)\;
          }
       }
    }
   }
  \Return{$P_1,P_2$}\;
 \caption{2p-greedy}
 \label{greedyalgo}
\end{algorithm}
\end{minipage}
\end{figure}

\paragraph{Second-Plane-Averse Plane Assignment based on Restriction Propagation on the Crossings Graph} While the greedy approach is very simple, it has the disadvantage that it may make suboptimal decisions leading to reduced coverage: assigning an arc to a given plane may seem like a good local decision, but depending on how arcs cross each other in the whole tree, it may lead to a subsequent situation where an arc cannot be assigned a plane even if the tree is actually 2-planar.

An example of this can be seen in Figure \ref{fig:2-greedy}: the greedy strategy will assign the arcs $w_1 \rightarrow w_3$ and $w_1 \rightarrow w_5$ to the first plane, which in a local context is the simplest thing to do. However, the fact that $w_1 \rightarrow w_3$ crosses $w_2 \rightarrow w_4$ (which is thus assigned to the second plane) and $w_3 \rightarrow w_6$ crosses both $w_1 \rightarrow w_5$ (first plane) and $w_2 \rightarrow w_4$ (second plane) then means that it is impossible to assign a plane to the arc $w_3 \rightarrow w_6$. This could have been prevented by assigning arc $w_1 \rightarrow w_5$ to the second plane, but a greedy algorithm has no way to anticipate this.
To deal with this problem, we propagate restrictions by traversing the crossings graph, i.e., a graph where its nodes represent the edges in the gold tree and two nodes are linked if the corresponding edges cross \cite{gomez-rodriguez-nivre-2013-divisible}.
Whenever we assign a given arc to plane 1, then we forbid plane 1 for its neighbors in the crossings graph (i.e. the arcs that cross it), we forbid plane 2 for the neighbors of its neighbors, plane 1 for the neighbors of those, and so on. For arcs assigned to plane 2, we proceed symmetrically.

Thus, the traversal order of arcs is the same as in the previous strategy, but for each new arc $a$, we look at the restrictions and assign it to the first plane if allowed, otherwise to the second plane if allowed, and finally to no plane if neither are allowed. In this case, the latter will only happen for non-2-planar trees: it is easy to show that situations where both planes are forbidden for the same arc can only happen if the crossings graph has a cycle of odd length, which is equivalent to the tree not being 2-planar (see \cite{gomez-rodriguez-nivre-2013-divisible}). Thus, this strategy guarantees full coverage of 2-planar structures. The pseudocode of the strategy can be seen in Algorithm~\ref{propagationalgo}, where $\overline{P_1}$ and $\overline{P_2}$ represent the arcs forbidden from planes 1 and 2, respectively.

\begin{figure}[ht]
  \centering
  \begin{minipage}{.7\linewidth}
 
    \begin{algorithm}[H]
\SetAlgoLined
  \KwIn{A set of arcs $T$, and input length $n$} 
  \KwResult{Two sets (planes) of arcs $P_1,P_2$}
\SetKwFunction{FPropagate}{Propagate}
\SetKwProg{Fn}{function}{:}{}

\Fn{\FPropagate{Edge sets $T,\overline{P_1},\overline{P_2}$, Edge $e$, Plane $i$}}{
  $\overline{P_i} \gets \overline{P_i} \cup \{ e \}$\; \tcp{ \normalfont $e$  forbidden from plane i}
  \For{($e' \in T \mid e'$ {\normalfont crosses} $e$)}{
    \uIf{$e' \not\in \overline{P_{3-i}}$}{
      $(\overline{P_1},\overline{P_2}) \gets$ \FPropagate{$T,\overline{P_1},\overline{P_2},e',3-i$}\;
    }
  }
  \Return{$\overline{P_1},\overline{P_2}$}\;
}

$P_1 \gets \varnothing$,
$P_2 \gets \varnothing$,
$\overline{P_1} \gets \varnothing$,
$\overline{P_2} \gets \varnothing$\;
 \For{$x_r \gets 1$ \textbf{to} $n$}{
    \For{$x_l \gets x_r-1$ \textbf{downto} $0$}{
       \If{$\exists a \in T \mid a = (x_l,x_r,l) \vee a = (x_r,x_l,l)$}{
         nextArc $\gets a$\;
         \uIf{{\normalfont nextArc} $\not\in \overline{P_1}$}{
              $P_1 \gets P_1 \cup \{ $nextArc$ \}$\;
              \FPropagate{$T,\overline{P_1},\overline{P_2},$nextArc,$2$}\;
           }
         \uElseIf{{\normalfont nextArc} $\not\in \overline{P_2}$}{
              $P_2 \gets P_2 \cup \{ $nextArc$ \}$\;
              \FPropagate{$T,\overline{P_1},\overline{P_2},$nextArc,$1$}\;
           }
          \uElse{
            do nothing (failed to assign nextArc to a plane)\;
          }
       }
    }
   }
  \Return{$P_1,P_2$}\;

 \caption{2p-prop}
 \label{propagationalgo}
\end{algorithm}
\end{minipage}
\end{figure}

\paragraph{Switch-averse plane assignment strategies} Another possibility is to implement variants of the previous two strategies that are \emph{switch-averse}, rather than \emph{second-plane-averse}. These variants work like the previous strategies, except for the difference than when both planes can be assigned to the current arc, we assign the last plane used, instead of always preferring to assign the first plane.

The implementation of the 2-planar transition-based parser by \newcite{gomez-rodriguez-nivre-2010-transition} used a switch-averse restriction-propagation strategy. This is a reasonable choice because in their transition-based parser it minimizes the number of transitions used: the algorithm's state holds the ``current'' plane being used, and switching to the other plane costs one transition.
In our sequence labeling context, where this is no longer true (the model always makes $n$ predictions for a sequence of length $n$), we made some initial experiments with switch-averse strategies but we found that they performed consistently (albeit slightly) worse than second-plane-averse strategies, so we discarded them for our experiments.

\begin{table}[]
\small{
\begin{adjustbox}{max width=1\linewidth}
\begin{tabular}{@{}lcc|lcc@{}}
\toprule
Language & \begin{tabular}[c]{@{}c@{}}\% non-projective\\  sentences\end{tabular} & \begin{tabular}[c]{@{}c@{}}\% non-projective\\  dependencies\end{tabular} & Language & \begin{tabular}[c]{@{}c@{}}\% non-projective\\ sentences\end{tabular} & \begin{tabular}[c]{@{}c@{}}\% non-projective\\ dependencies\end{tabular} \\ \midrule
Ancient Greek\textsubscript{Perseus} & 63.87 & 10.14 & Korean\textsubscript{Kaist} & 21.70 & 2.55 \\
Basque\textsubscript{BDT} & 33.17 & 4.69 &Danish\textsubscript{DDT} & 21.50 & 1.74 \\
Hungarian\textsubscript{Szeged} & 27.11 & 1.97 & Gothic\textsubscript{PROIEL} & 17.57 & 2.53 \\
Portuguese\textsubscript{Bosque} & 23.31 & 1.85 & Lithuanian\textsubscript{HSE} & 17.49 & 1.27 \\
Urdu\textsubscript{UDTB}  & 22.57 & 1.32 & \textit{Japanese\textsubscript{GSD}} & \textit{0} & \textit{0} \\
Afrikaans\textsubscript{AfriBooms} & 22.34 & 1.62 &\textit{Galician\textsubscript{CTG}}& \textit{0} & \textit{0} \\ \bottomrule
\end{tabular}
\end{adjustbox}}
\caption{Percentage of non-projective sentences and dependencies of the selected UD treebanks, where Japanese\textsubscript{GSD} and Galician\textsubscript{CTG} are control treebanks.}
\label{tab:ratio_non-proj}
\end{table}

\section{Bracketing decoding}
\label{sec:decoding}

When a sentence is represented with the bracketing encoding in a single plane, a valid left arc is associated with a pair of matching brackets $\lT$ and $\rBS$ while a right arc is associated with a pair of $\lSL$ and $\rT$. For each sentence we create two initially empty stacks, $\sigma_L$ and $\sigma_R$, in order to keep the elements separate with respect to the arc direction. Thus, the output labels generated by the system are read from left to right, decomposed into their brackets, and then brackets corresponding to left arcs are processed in $\sigma_L$ and those that encode right arcs are processed in $\sigma_R$. In order to handle a second plane with brackets represented as ($\lT^{*}$, $\rBS^{*}$) and ($\lSL^{*}$, $\rT^{*}$), we simply use additional stacks: $\sigma_L^{*}$ and $\sigma_R^{*}$.

More particularly, decoding proceeds by reading a label for each token and pushing each opening bracketing element to the corresponding stack while preserving the token's index. For instance, when reading a new label that contains $\lT$, the bracket element is pushed into the $\sigma_L$ stack and can only be popped once there is a later matching label with a closing bracketing element $\rBS$ that will be used to create a left arc, by recovering the index stored together with the $\lT$ bracket. Analogously, right arcs are processed in the same way, but in a different stack.

\paragraph{Postprocessing}

Decoded labels do not ensure creating a well-formed tree. For that reason, we adapt some common heuristics for all encodings in order to postprocess them. In case some of the brackets in any of the stacks are unbalanced, the outermost bracket elements are discarded. Tokens that are not assigned any head are recovered by attaching them to the word that is attached to the dummy root (i.e., the syntactic head of the sentence). Cycles are also solved by removing the leftmost arc in the cycle.

\section{Experiments}
\paragraph{Data}
We extracted the most non-projective treebanks from UDv2.4 \cite{ud2.4} based on the percentage of non-projective sentences, and discarded some of them due to the lack of a pre-trained UDpipe model or due to the lack of a development set. The selected treebanks were: Ancient Greek\textsubscript{Perseus},
Basque\textsubscript{BDT},
Hungarian\textsubscript{Szeged},
Portuguese\textsubscript{Bosque},
Urdu\textsubscript{UDTB},
Afrikaans\textsubscript{AfriBooms},
Korean\textsubscript{Kaist},
Danish\textsubscript{DDT},
Gothic\textsubscript{PROIEL},
Lithuanian\textsubscript{HSE}.
In addition, two fully projective treebanks (Galician\textsubscript{CTG} and Japanese\textsubscript{GSD}) were included as control treebanks. Table \ref{tab:ratio_non-proj} shows the selected treebanks with their percentage of non-projective sentences and dependencies.
For all of them, we ran UDPipe models \cite{straka-strakova-2017-tokenizing} to obtain predicted segmentation and tokenization. 
We also computed predicted PoS tags, but they were not used (nor gold PoS tags were) to train \emph{any} of the models, but just to decode the labels from the rel-PoS encoding \cite{strzyz-etal-2019-viable}. In addition, we included dummy beginning- and end-of-sentence tokens (\textsc{bos},\textsc{eos}) as in previous work in parsing as labeling.

\paragraph{Model}

For our experiments we use bidirectional long short-term memory networks \cite{hochreiter1997long,schuster1997bidirectional} as implemented in the NCRF++ framework \cite{yang2018ncrf++}.\footnote{We omit however the CRF on top of the BiLSTMs.} Each input word $w_i$ is represented as a vector which comes from a concatenation of (i) an external pre-trained word embedding,  which is further fine-tuned during training,
and (ii) a second word embedding which results from the output of a char-LSTM, which is trained end-to-end together with the rest of the network.

In this context, let LSTM$_\theta(\vec{x})$ be a black-box long short-term memory network that processes the sequence of vectors $\vec{x}=[\vec{x}_1,...,\vec{x}_{|\vec{x}|}]$, then the output for $\vec{x}_i$ is a hidden vector $\vec{h_i}$ which represents the word based on its left and right sentence context:

\begin{equation}
\vec{h}_i = \text{BiLSTM}_\theta(\vec{x},i) = \text{LSTM}_\theta^l(\vec{x}_{[1:i]}) \circ \text{LSTM}_\theta^r(\vec{x}_{[|\vec{x}|:i]}).
\end{equation}

More particularly, we stack 2 BiLSTMs before computing the output layer. For this, we consider a simple hard-sharing multi-task learning architecture, where each $\vec{h_i}$ is sent to three separate layers in order to generate the classifications through regular softmaxes: two labels predicted for each plane (one label per plane)\footnote{If a given word has no arcs associated to that plane}, we generate an empty label $\emptyset$. and another one for the word's dependency relation. Afterwards, label decoding is followed by a postprocessing step with some heuristics to ensure a valid dependency tree (as described in \S \ref{sec:decoding}).

\subsection{Analysis and results}

Next, we compare the performance of the original bracketing encoding (1p-brackets), 2-planar with greedy plane assignment (2p-greedy) and 2-planar with restriction propagation (2p-prop) with respect to their theoretical arc coverage, as well as their empirical recall and precision. 
For UAS/LAS, we also report results for models trained on the rel-PoS encoding.

\begin{table}[hptb!]
\centering
\small{
\renewcommand{\arraystretch}{1.1}
\begin{tabular}{@{}lccc|lccc@{}}
\toprule
Language & 1p-brackets & 2p-greedy & 2p-prop & Language & \multicolumn{1}{c}{1p-brackets} & \multicolumn{1}{c}{2p-greedy} &2p-prop  \\ \midrule
Ancient Greek\textsubscript{Perseus} & 89.53 &99.27 & \textbf{99.33}  & Afrikaans\textsubscript{AfriBooms} & 98.65 & \textbf{99.99}   &\textbf{99.99}  \\
Basque\textsubscript{BDT} & 94.85 & \textbf{99.85}  & 99.62 & Korean\textsubscript{Kaist} &  98.42 & \textbf{100.00} & \textbf{100.00}  \\
Hungarian\textsubscript{Szeged} & 97.57 & 99.96  &\textbf{99.98}  & Danish\textsubscript{DDT} & 98.10 & \textbf{99.97} &99.96   \\
Portuguese\textsubscript{Bosque} & 98.10 & \textbf{99.95}  & 99.88 & Gothic\textsubscript{PROIEL} & 97.58 & 99.94 &\textbf{99.98}   \\
Urdu\textsubscript{UDTB}&  98.68 & \textbf{99.95}  &99.94   & Lithuanian\textsubscript{HSE} &  98.35 & 99.97 &\textbf{100.00}  \\ \bottomrule
\end{tabular}}

\caption{Percentage of arcs covered by the proposed encodings on the gold training set from highly non-projective treebanks.}
\label{tab:sanity-check}
\end{table}

\paragraph{Theoretical advantage} Table \ref{tab:sanity-check} compares the dependency arc coverage by the encodings on the gold training sets. It is easy to conclude that the 2-planar encodings almost fully succeed to reconstruct highly non-projective datasets, while the bracketing encoding suffers more. When comparing both plane assignments for the 2-planar encodings, we see that the coverage of 2p-greedy is already so high (99.9\% or more in all but two treebanks) that the extra coverage provided by 2p-prop is not large in absolute terms. In fact, in some treebanks, 2p-prop even has slightly less measured coverage than 2p-greedy, even though (as explained earlier) the former guarantees full coverage of 2-planar trees while the latter does not. This can be explained because there are \emph{non-2-planar} trees where 2p-greedy happens to cover more arcs. In such trees, the theoretical guarantee provided by 2p-prop does not apply.

\begin{table}[hptb!]
\centering
\begin{adjustbox}{max width=1\linewidth}
\begin{tabular}{@{}clcccclcccclccc@{}}
\toprule
Language & Encoding & \begin{tabular}[c]{@{}c@{}}Task 1\\ (1st plane)\end{tabular} & \begin{tabular}[c]{@{}c@{}}Task 2\\ (2nd plane)\end{tabular} & \begin{tabular}[c]{@{}c@{}}Task 3\\ (deprel)\end{tabular} & Language & Encoding & \begin{tabular}[c]{@{}c@{}}Task 1\\ (1st plane)\end{tabular} & \begin{tabular}[c]{@{}c@{}}Task 2\\ (2nd plane)\end{tabular} & \begin{tabular}[c]{@{}c@{}}Task 3\\ (deprel)\end{tabular} & Language & Encoding & \begin{tabular}[c]{@{}c@{}}Task 1\\ (1st plane)\end{tabular} & \begin{tabular}[c]{@{}c@{}}Task 2\\ (2nd plane)\end{tabular} & \begin{tabular}[c]{@{}c@{}}Task 3\\ (deprel)\end{tabular} \\ \midrule
\multirow{4}{*}{\begin{tabular}[c]{@{}c@{}}Ancient\\ Greek\end{tabular}} & rel-PoS & 166 & -- & \multicolumn{1}{c|}{27} & \multirow{4}{*}{Urdu} & rel-PoS & 190 & -- & \multicolumn{1}{c|}{27} & \multirow{4}{*}{Gothic} & rel-PoS & 121 & -- & 34 \\
 &1p-brackets & 210 & -- & \multicolumn{1}{c|}{27} &  &1p-brackets& 95 & -- & \multicolumn{1}{c|}{27} &  & 1p-brackets & 114 & -- & 34 \\
 & 2p-greedy & 108 & 37 & \multicolumn{1}{c|}{27} &  & 2p-greedy & 80	&22& \multicolumn{1}{c|}{27} &  & 2p-greedy &78	&18 & 34 \\
 & 2p-prop & 109 & 39 & \multicolumn{1}{c|}{27} &  & 2p-prop & 80&	22& \multicolumn{1}{c|}{27} &  & 2p-prop& 78&	19 & 34 \\ \midrule
\multirow{4}{*}{Basque} & rel-PoS & 132 & -- & \multicolumn{1}{c|}{32} & \multirow{4}{*}{Afrikaans} & rel-PoS & 110 & -- & \multicolumn{1}{c|}{28} & \multirow{4}{*}{Lithuanian} & rel-PoS & 89 & -- & 38 \\
 & 1p-brackets & 134 & -- & \multicolumn{1}{c|}{32} &  & 1p-brackets & 77 & -- & \multicolumn{1}{c|}{28} &  & 1p-brackets & 57 & -- & 38 \\
 & 2p-greedy & 84&25 & \multicolumn{1}{c|}{32} &  & 2p-greedy & 62	&15& \multicolumn{1}{c|}{28} &  &2p-greedy & 46	&11 & 38 \\
 & 2p-prop & 83&25 & \multicolumn{1}{c|}{32} &  & 2p-prop & 62 & 15 & \multicolumn{1}{c|}{28} &  & 2p-prop & 46&	12 & 38 \\ \midrule
\multirow{4}{*}{Hungarian} & rel-PoS & 128 & -- & \multicolumn{1}{c|}{56} & \multirow{4}{*}{Korean} & rel-PoS & 134 & -- & \multicolumn{1}{c|}{32} & \multirow{4}{*}{\textit{Japanese}} & \textit{rel-PoS} & \textit{77} & \textit{--} & \textit{27} \\
 &1p-brackets & 101 & -- & \multicolumn{1}{c|}{56} &  & 1p-brackets & 89 & -- & \multicolumn{1}{c|}{32} &  & \textit{1p-brackets} & \textit{45} & \textit{--} & \textit{27} \\
 & 2p-greedy & 71	&19 & \multicolumn{1}{c|}{56} &  & 2p-greedy & 73&	14 & \multicolumn{1}{c|}{32} &  & \textit{2p-greedy} & \textit{45} & \textit{3} & \textit{27} \\
 & 2p-prop & 71&	21 & \multicolumn{1}{c|}{56} &  & 2p-prop & 73 & 14 & \multicolumn{1}{c|}{32} &  & \textit{2p-prop} & \textit{45} & \textit{3} & \textit{27} \\ \midrule
\multirow{4}{*}{Portuguese} & rel-PoS & 192 & -- & \multicolumn{1}{c|}{43} & \multirow{4}{*}{Danish} & rel-PoS & 150 & -- & \multicolumn{1}{c|}{38} & \multirow{4}{*}{\textit{Galician}} & \textit{rel-PoS} & \textit{132} & \textit{--} & \textit{26} \\
 & 1p-brackets & 110 & -- & \multicolumn{1}{c|}{43} &  & 1p-brackets & 128 & -- & \multicolumn{1}{c|}{38} &  & \textit{1p-brackets} & \textit{82} & \textit{--} & \textit{26} \\
 & 2p-greedy & 88&	25 & \multicolumn{1}{c|}{43} &  & 2p-greedy & 97&	23 & \multicolumn{1}{c|}{38} &  & \textit{2p-greedy} & \textit{82} & \textit{3} & \textit{26} \\
 & 2p-prop& 88	&27 & \multicolumn{1}{c|}{43} &  &2p-prop & 96&	25 & \multicolumn{1}{c|}{38} &  & \textit{2p-prop} & \textit{82} & \textit{3} & \textit{26} \\ \bottomrule
\end{tabular}
\end{adjustbox}
\caption{Label size of each encoding based on the training and dev set for each treebank. Each task contains three additional labels: \textsc{bos}, \textsc{eos} and \textsc{$\emptyset$}. Hence the Japanese and Galician treebanks have three labels for the second plane, although they are fully projective.}
\label{tab:label_size}
\end{table}

With respect to the number of labels that each encoding generates (which will directly impact the output size of the softmax layers), Table \ref{tab:label_size} shows the comparison of the output vocabulary sets for each of the tasks in the multi-task learning setup. We can see that, for most languages, bracketing encodings generate a smaller tag set than rel-PoS\footnote{In the rel-PoS encoding, each label word represents the head based on a PoS-tag offset. See also \cite{strzyz-etal-2019-viable}.}; and in general, the 2-planar encodings do not produce increases in tagset size with respect to the 1-planar bracketing encoding. In fact, for the most non-projective languages (like Ancient Greek or Basque), the 2-planar encodings clearly compress the tag set as, in spite of having a larger variety of brackets, they appear distributed among the two planes so that the bracket strings in each label will tend to be shorter.

\begin{table}[htpb!]
\centering
\begin{adjustbox}{max width=1\linewidth}
\renewcommand{\arraystretch}{1}
\setlength{\tabcolsep}{12pt}
\small{
\begin{tabular}{@{}lcccccc@{}}
\toprule
\multirow{2}{*}{Language} & \multicolumn{2}{c}{1p-brackets} & \multicolumn{2}{c}{\begin{tabular}[c]{@{}c@{}}2p-greedy\end{tabular}} & \multicolumn{2}{c}{\begin{tabular}[c]{@{}c@{}}2p-prop\end{tabular}} \\
 & P & R & P & R & P & R \\ \midrule
Ancient Greek\textsubscript{Perseus}& 85.74 &	54.34 & 86.33	&63.85 &\bf 87.58	&\bf 66.23	 \\
Basque\textsubscript{BDT} & 69.87&	45.80& 70.14&	\bf 52.97&\bf 72.77&	52.80   \\
Hungarian\textsubscript{Szeged} &37.17&	66.98& 35.51&	71.70&\bf 37.80&\bf	74.53  \\
Portuguese\textsubscript{Bosque} & 52.94&	24.77& 55.84	&39.45&\bf 61.64&\bf	41.28  \\
Urdu\textsubscript{UDTB} & 36.63&	\bf36.63& 38.10&	31.68&\bf 39.78&\bf	36.63  \\
Afrikaans\textsubscript{AfriBooms} &40.99& 65.35&46.72	&63.37&\bf 46.94&\bf	68.32 \\
Korean\textsubscript{Kaist} & 59.45&\bf 	49.54&62.24&	47.03&\bf 62.80&	47.03 \\
Danish\textsubscript{DDT} & 45.54&	\bf 48.57&\bf 46.36&	\bf 48.57&45.37&	46.67 \\
Gothic\textsubscript{PROIEL} & 50.50&	26.42&\bf 58.88&	32.64&56.00&\bf	36.27 \\
Lithuanian\textsubscript{HSE} & \bf 34.38&	\bf 91.67& 27.59	&66.67&33.33&	83.33 \\ \midrule
\textit{Average} & 51.32&	51.01&	52.77&	51.79&	\bf 54.40& \bf	55.31  \\\bottomrule
\end{tabular}}
\end{adjustbox}
\caption{Models' precision and recall of non-projective sentences on the test set.}
\label{tab:recall-precision-sent}
\end{table}

\begin{table}[htpb!]
\centering
\begin{adjustbox}{max width=1\linewidth}
\renewcommand{\arraystretch}{1}
\setlength{\tabcolsep}{12pt}
\small{
\begin{tabular}{@{}lcccccc@{}}
\toprule
\multirow{2}{*}{Language} & \multicolumn{2}{c}{1p-brackets} & \multicolumn{2}{c}{\begin{tabular}[c]{@{}c@{}}2p-greedy\end{tabular}} & \multicolumn{2}{c}{\begin{tabular}[c]{@{}c@{}}2p-prop\end{tabular}} \\
 & P & R & P & R & P & R \\ \midrule
Ancient Greek\textsubscript{Perseus}& 20.82&	10.32&\bf	32.40&	18.65&31.40&	\bf 19.16 \\
Basque\textsubscript{BDT} & 18.41&	11.80&	28.11&	19.83&\bf 31.76&\bf	20.40  \\
Hungarian\textsubscript{Szeged} & 1.57&	4.05&3.13	&9.25&\bf 4.06&\bf	10.98  \\
Portuguese\textsubscript{Bosque} & 10.87&	5.18&14.50	&9.84&\bf 20.18&\bf	11.92 \\
Urdu\textsubscript{UDTB} & 3.26&	\bf 3.92&	 0.69&	0.65& \bf3.41&	\bf 3.92   \\
Afrikaans\textsubscript{AfriBooms} & 11.04&	18.09&\bf	13.09	&\bf 19.15&12.59&	18.62  \\
Korean\textsubscript{Kaist} & 28.12&	\bf21.68&\bf	32.26&	21.37&31.06&	21.53 \\
Danish\textsubscript{DDT} &7.66&	11.35&	\bf 12.96&	\bf 19.86&9.45&	13.48 \\
Gothic\textsubscript{PROIEL} & 11.11&	5.17&\bf 19.63&	11.03&17.46&\bf	11.38 \\
Lithuanian\textsubscript{HSE} & 0&	0&	0&	0&	\textbf{1.64}&	\textbf{6.25}  \\ \midrule
\textit{Average} &11.29&	9.16&	15.68&	12.96& \bf	16.30& \bf	13.76   \\\bottomrule
\end{tabular}}
\end{adjustbox}
\caption{Models' precision and recall of non-projective dependencies on the test set. }
\label{tab:recall-precision}
\end{table}

\paragraph{Results} To investigate how the coverage in Table \ref{tab:sanity-check} translates into non-projective performance in actual parsing,
we report models' precision and recall. In Table \ref{tab:recall-precision-sent}, the precision and recall on \textit{non-projective sentences}\footnote{Precision and recall on non-projective sentences are computed by looking whether a given sentence is identified as non-projective (i.e. given a non-projective parse), disregarding the correctness of the predicted non-projective dependencies for that sentence.} increase across the treebanks with 2-planar models, suggesting that they are capable of identifying non-projective sentences to a greater extent than the original bracketing model. Table \ref{tab:recall-precision} shows that 2p-greedy and 2p-prop models improve the recall and precision of \textit{non-projective dependencies} in the majority of treebanks.\footnote{The reported precision and recall for Lithuanian is lower than for the other treebanks. As we show in Appendix \ref{appendix-treebank-sizes}, the Lithuanian treebank contains a small number of sentences and therefore it is hard to draw robust conclusions about its poor performance.} Again, 2-planar encodings outperform the original bracketing baseline, even though the latter is able to cover non-projectivity to some degree (crossing arcs pointing in opposite directions). Both 2p-greedy and 2p-prop obtain similar scores, showing that their coverage is comparable.

\begin{table}[hptb!]
\centering
\begin{adjustbox}{max width=1\linewidth}
\begin{tabular}{@{}clllll|clllll@{}}
\toprule
\multicolumn{1}{l}{\multirow{2}{*}{Language}} & \multirow{2}{*}{Encoding} & \multicolumn{2}{c}{dev} & \multicolumn{2}{c}{test} &\multicolumn{1}{l}{\multirow{2}{*}{Language}} & \multirow{2}{*}{Encoding} & \multicolumn{2}{c}{dev} & \multicolumn{2}{c}{test} \\
\multicolumn{1}{l}{} &  & \multicolumn{1}{c}{UAS} & \multicolumn{1}{c}{LAS} & \multicolumn{1}{c}{UAS} & \multicolumn{1}{c}{LAS} & \multicolumn{1}{l}{} &  & \multicolumn{1}{c}{UAS} & \multicolumn{1}{c}{LAS} & \multicolumn{1}{c}{UAS} & \multicolumn{1}{c}{LAS}\\ \midrule
\multirow{4}{*}{\begin{tabular}[c]{@{}c@{}}Ancient \\ Greek\textsubscript{Perseus}\end{tabular}} 
& rel-PoS & 65.29&	58.27	&62.91&	55.07 & \multirow{4}{*}{\begin{tabular}[c]{@{}c@{}}Korean\textsubscript{Kaist} \end{tabular}} 
& rel-PoS & 81.47&	78.50&	77.25&	73.92\\ \cmidrule(l){2-6} \cmidrule(l){8-12} 
 & 1p-brackets & 64.70&	57.21	&63.36&	54.80 & 
 & 1p-brackets & 84.54&	81.54&	\textbf{82.37}&	\textbf{79.03} \\
 & 2p-greedy &  67.10 &	59.97& \textbf{65.90} &	\textbf{57.15}	 & 
 &2p-greedy &85.01&	82.01&82.33&	78.91  \\
 & 2p-prop & 67.06&	59.84&65.11&	56.55 &
 &2p-prop & 84.65	&81.73&82.32&	\textbf{79.03} \\ \midrule
\multirow{4}{*}{Basque\textsubscript{BDT}} & 
rel-PoS & 77.48	&72.91&	75.28&	70.19 & \multirow{4}{*}{Danish\textsubscript{DDT}} 
& rel-PoS & 78.28&	74.93&	77.07&	73.45 \\ \cmidrule(l){2-6} \cmidrule(l){8-12} 
 & 1p-brackets & 80.13	&75.37&	78.37&	72.95 & 
 & 1p-brackets & 80.60&	76.59&	78.25&	73.94 \\
 & 2p-greedy & 79.98 &	75.18&78.13&	72.63 & 
 &2p-greedy & 80.68&	76.80&78.49&	74.07 \\
 & 2p-prop &   	80.44	&75.56&\textbf{78.58}&	\textbf{73.08}&
 &2p-prop &  	81.15	&77.27&\textbf{78.87}&	\textbf{74.42} \\ \midrule
\multirow{4}{*}{Hungarian\textsubscript{Szeged}} & rel-PoS & 72.58&	67.13&	66.19&	59.32
& \multirow{4}{*}{Gothic\textsubscript{PROIEL}} 
& rel-PoS & 65.25&	58.58& \bf	67.14&	\bf 59.72 \\ \cmidrule(l){2-6} \cmidrule(l){8-12}
 & 1p-brackets & 75.09&	69.13&	67.80&	60.50 & 
 & 1p-brackets & 65.26&	57.92&	66.63&	59.02 \\
 & 2p-greedy & 75.47&	69.33&\textbf{68.07}&	\textbf{60.74} & 
 &2p-greedy & 65.26&	58.05&66.84&	59.26\\
 & 2p-prop &	75.26&	69.05&67.95&	60.63 &
 &2p-prop & 	65.29	&57.91&66.25&	58.41 \\ \midrule
\multirow{4}{*}{Portuguese\textsubscript{Bosque}} & rel-PoS & 87.10&	84.28&	84.74&	81.02 & \multirow{4}{*}{Lithuanian\textsubscript{HSE}} & rel-PoS & 39.04&	26.37&	31.05&	19.70 \\ \cmidrule(l){2-6} \cmidrule(l){8-12}
 & 1p-brackets & 88.88&	85.78&	\textbf{86.67}&	\textbf{82.44}& 
 & 1p-brackets & 40.97&	25.63&	34.62&	19.42 \\
 & 2p-greedy & 88.88&	85.76&86.51&	82.39 & 
 &2p-greedy & 41.34&	26.46&\textbf{35.08}&	20.45 \\
 & 2p-prop& 	89.00&	85.82&86.52&	82.17&
 &2p-prop & 	44.19	&29.03&34.80&	\textbf{21.29} \\ \midrule 
\multirow{4}{*}{Urdu\textsubscript{UDTB}} & rel-PoS & 80.98&	75.09&	81.18&	75.26 & \multirow{4}{*}{\textit{Japanese\textsubscript{GSD}}} &
\textit{rel-PoS} & \textit{76.60}&	\textit{75.83}&	\textit{74.83}&	\textit{73.96} \\ \cmidrule(l){2-6} \cmidrule(l){8-12} 
 & 1p-brackets & 84.22&	77.23&	\textbf{84.28}&	77.19 & 
 & \textit{1p-brackets} & \textit{78.73}&	\textit{77.67}&	\textit{77.34}&	\textit{76.10} \\ 
 & 2p-greedy & 84.01&	77.16&84.08&	77.19&
 & \textit{2p-greedy} &  \textit{78.81}&	\textit{77.78}&\textit{77.47}&	\textit{76.24}\\
 & 2p-prop &	83.89&	77.30&84.26&	\textbf{77.41} &
 &\textit{2p-prop} & 	\textit{78.81}	&\textit{77.78}&\textit{77.47}&	\textit{76.24}  \\ \midrule 
\multirow{4}{*}{Afrikaans\textsubscript{AfriBooms}} & rel-PoS & 79.00&	74.58&	78.93&	74.65  & \multirow{4}{*}{\textit{Galician\textsubscript{CTG}}} & 
\textit{rel-PoS} & \textit{79.72}&	\textit{76.40}&	\textit{78.36}&	\textit{75.05}  \\ \cmidrule(l){2-6} \cmidrule(l){8-12}
 & 1p-brackets & 80.77&	75.54&	79.52&	74.86 & 
 & \textit{1p-brackets} & \textit{80.82}&	\textit{77.35}&	\textit{80.02}&	\textit{76.33} \\
 & 2p-greedy & 81.41&	76.33&\textbf{80.13}&	\textbf{75.53} & 
 &\textit{2p-greedy} & \textit{80.90}&	\textit{77.36}&\textit{79.91}&	\textit{76.32} \\
 & 2p-prop & 81.50&	76.30&79.96&	75.43 &
 &\textit{2p-prop} & \textit{80.90}&	\textit{77.36}&\textit{79.91}&	\textit{76.32}\\ \bottomrule
\end{tabular}
\end{adjustbox}
\caption{UAS and LAS (\%) for the respective encodings on the predicted dev and test set of highly non-projective treebanks and control treebanks. 
}
\label{tab:final}
\end{table}

Table \ref{tab:final} compares the LAS and UAS performance of the 1- and 2-planar, and also of the rel-PoS encoding.\footnote{Note that the implementations compared here do not use PoS tags as features. This is sensible for the bracketing encodings, and for the focus of this paper where we are interested in encodings that can be run using raw words as the only input, but differs from the standard setup in the rel-PoS encoding (where using PoS tags comes at no extra significant cost, because they need to be computed for decoding in any case).} 2-planar encodings outperform the 
existing bracketing
encoding in the majority of treebanks. The gains vary between languages but on average 2p-greedy improves UAS by $0.4$ and 2p-prop by $0.3$, and both improve LAS by $0.4$ across highly non-projective treebanks. Comparing both assignment strategies for the 2-planar encoding, the theoretical advantage in coverage provided by 2p-prop over 2p-greedy 
does not translate into accuracy gains in general, 
as the actual difference in coverage is small when measured in the treebanks (as was seen in Table \ref{tab:sanity-check}) and
the simpler greedy assignment strategy is likely to be easier to learn by the machine learning setup. 

Since the syntactic dependencies are represented by a finite set of labels that have been seen in the training  and development sets,
as in all parsing as sequence labeling approaches,
it is expected that at test time our model may encounter unseen labels. In Appendix \ref{appendix-label-coverage} we show the label coverage of all encodings on the test set. In general, it seems that the unseen labels do not have significant impact on the overall performance due to their rare occurrence.

\begin{table}[hptb!]
\centering
\begin{adjustbox}{max width=1\linewidth}
\begin{tabular}{@{}llccllccllcc@{}}
\toprule
\multirow{2}{*}{Language} & \multirow{2}{*}{Encoding} & \multicolumn{2}{c}{sent/s} & \multirow{2}{*}{Language} & \multirow{2}{*}{Encoding} & \multicolumn{2}{c}{sent/s} & \multirow{2}{*}{Language} & \multirow{2}{*}{Encoding} & \multicolumn{2}{c}{sent/s} \\
 &  & CPU & GPU &  &  & CPU & GPU &  &  & CPU & GPU \\ \midrule
\multirow{4}{*}{\begin{tabular}[c]{@{}l@{}}Ancient \\ Greek\textsubscript{Perseus}\end{tabular}} & rel-PoS & 305* & \multicolumn{1}{c|}{1012*} & \multirow{4}{*}{Urdu\textsubscript{UDTB}} & rel-PoS & 182* & \multicolumn{1}{c|}{625*} & \multirow{4}{*}{Gothic\textsubscript{PROIEL}} & rel-PoS & 430* & 1266* \\
 & 1p-brackets & 303 & \multicolumn{1}{c|}{1011} &  & 1p-brackets & 186 & \multicolumn{1}{c|}{616} &  & 1p-brackets & 429& 1269 \\
 & 2p-greedy & 288 & \multicolumn{1}{c|}{889} &  & 2p-greedy & 174 & \multicolumn{1}{c|}{544} &  & 2p-greedy & 414 & 1175 \\
 & 2p-prop & 289 & \multicolumn{1}{c|}{886} &  & 2p-prop & 175 & \multicolumn{1}{c|}{549} &  & 2p-prop & 412 & 1186 \\ \midrule
\multirow{4}{*}{Basque\textsubscript{BDT}} & rel-PoS & 387* & \multicolumn{1}{c|}{1461*} & \multirow{4}{*}{Afrikaans\textsubscript{AfriBooms}} & rel-PoS & 228* & \multicolumn{1}{c|}{861*} & \multirow{4}{*}{Lithuanian\textsubscript{HSE}} & rel-PoS & 239* & 828* \\
 & 1p-brackets & 388 & \multicolumn{1}{c|}{1454} &  & 1p-brackets & 228 & \multicolumn{1}{c|}{857} &  & 1p-brackets & 235 & 769 \\
 & 2p-greedy & 378 & \multicolumn{1}{c|}{1369} &  & 2p-greedy & 220 & \multicolumn{1}{c|}{805} &  & 2p-greedy & 228 & 740 \\
 & 2p-prop & 378 & \multicolumn{1}{c|}{1369} &  & 2p-prop & 221 & \multicolumn{1}{c|}{805} &  & 2p-prop & 229 & 730 \\ \midrule
\multirow{4}{*}{Hungarian\textsubscript{Szeged}} & rel-PoS & 219* & \multicolumn{1}{c|}{802*} & \multirow{4}{*}{Korean\textsubscript{Kaist}} & rel-PoS & 442* & \multicolumn{1}{c|}{1718*} & \multirow{4}{*}{\textit{Japanese\textsubscript{GSD}}} & \textit{rel-PoS} & \textit{214*} & \textit{663*} \\
 & 1p-brackets & 221 & \multicolumn{1}{c|}{797} &  & 1p-brackets & 447 & \multicolumn{1}{c|}{1718} &  & \textit{1p-brackets} & \textit{214} & \textit{661} \\
 & 2p-greedy & 212 & \multicolumn{1}{c|}{739} &  & 2p-greedy & 434 & \multicolumn{1}{c|}{1598} &  & \textit{2p-greedy} & \textit{206} & \textit{611} \\
 & 2p-prop & 213 & \multicolumn{1}{c|}{750} &  & 2p-prop & 435 & \multicolumn{1}{c|}{1544} &  & \textit{2p-prop} & \textit{205} & \textit{616} \\ \midrule
\multirow{4}{*}{Portuguese\textsubscript{Bosque}} & rel-PoS & 242* & \multicolumn{1}{c|}{868*} & \multirow{4}{*}{Danish\textsubscript{DDT}} & rel-PoS & 279* & \multicolumn{1}{c|}{962*} & \multirow{4}{*}{\textit{Galician\textsubscript{CTG}}} & \textit{rel-PoS} & \textit{175*} & \textit{752*} \\
 & 1p-brackets & 246 & \multicolumn{1}{c|}{872} &  & 1p-brackets & 280 & \multicolumn{1}{c|}{980} &  & \textit{1p-brackets} & \textit{177} & \textit{756} \\
 & 2p-greedy & 236 & \multicolumn{1}{c|}{811} &  & 2p-greedy & 265 & \multicolumn{1}{c|}{841} &  & \textit{2p-greedy} & \textit{170} & \textit{673} \\
 & 2p-prop & 237 & \multicolumn{1}{c|}{814} &  & 2p-prop & 264 & \multicolumn{1}{c|}{832} &  & \textit{2p-prop} & \textit{170} & \textit{673} \\ \bottomrule
\end{tabular}
\end{adjustbox}
\caption{Comparison of parsing speeds (sent/s) on a single core CPU and GPU. The reported speeds are averaged over 5 runs. 
Times with a * are a reminder that the rel-PoS encoding additionally requires PoS tagging, whose time is not included in these speeds. As an example, the UDPipe tagging speed on the test set (in sent/s) is: Ancient Greek-$567$, Basque-$881$, Hungarian-$755$, Portuguese-$210$, Urdu-$45$, Afrikaans-$465$, Korean-$921$, Danish-$671$, Gothic-$861$, Lithuanian-$1418$, Japanese-$767$ and Galician-$366$.
}
\label{tab:speeds}
\end{table}

Finally, we measured speeds for each of the encodings on various treebanks, run on a single core CPU\footnote{For CPU experiments, we used a CPU core Intel Core i7-8700 CPU 3.2 GHz.} and GPU\footnote{For GPU experiments, we used an Nvidia TITAN Xp.}, which we breakdown in Table \ref{tab:speeds}. We can observe that the speed is very similar between 1-planar and 2-planar encodings. This is because the bottleneck of the model is in the BiLSTMs, and computing the softmaxes comes at almost no cost despite the differences in the output vocabularies.

\section{Conclusion}

We have shown a new bracketing-based linearization of 2-planar trees compatible with parsing as sequence labeling.
Our main goal was to introduce a bracketing encoding with the ability to perform unrestricted non-projective dependency parsing, which remained as an open challenge in sequence labeling parsing under the family of bracketing encodings. Together with the proposed plane assignment strategies and a BiLSTM-based network, our 2-planar bracket representations improve the performance over the existing bracketing-based encoding for parsing as sequence labeling, and also outperform the PoS-based encoding in the absence of PoS-tags as input parameters to the model.
Thus, it can be a useful alternative where an encoding that depends on PoS tags is not desirable, e.g. domains with low-frequency or low-quality PoS tags, or to decrease even further the latency of sequence labeling parsers.

Finally, it is worth noting that we have proposed plane assignment strategies that minimize the use of the second plane. However, it is a possible avenue for future work to examine other strategies based on different criteria than the one presented in this paper.

\section*{Acknowledgements}

This work has received funding from the European Research Council (ERC), which has funded this research under the European Union's Horizon 2020 research and innovation programme (FASTPARSE, grant agreement No 714150), from MINECO (ANSWER-ASAP, TIN2017-85160-C2-1-R), from Xunta de Galicia (ED431C 2020/11), and from Centro de Investigación de Galicia `CITIC', funded by Xunta de Galicia and the European Union (European Regional Development Fund- Galicia 2014-2020 Program), by grant ED431G 2019/01. DV is supported by a 2020 Leonardo Grant for Researchers and Cultural Creators from the BBVA Foundation.

\bibliographystyle{coling}
\bibliography{coling2020}

\clearpage
\appendix

\section{Treebank sizes}\label{treebanks-stats} \label{appendix-treebank-sizes}

We provide some statistics about the chosen treebanks. In Table \ref{tab:treebank-size}, we report the total number of sentences for each dataset split with their respective non-projectivity percentage.

\begin{table}[hptb!]
\centering
\small{
\begin{adjustbox}{max width=1\linewidth}
\begin{tabular}{@{}lrrr|lrrr@{}}
\toprule
Language & Train & Dev & Test & Language & \multicolumn{1}{c}{Train} & \multicolumn{1}{c}{Dev} &Test  \\ \midrule
Ancient Greek\textsubscript{Perseus} & 11476 (\textit{62.77\%}) & 1137 (\textit{74.41\%})& 1306 (\textit{64.40\%})
& Afrikaans\textsubscript{AfriBooms} & 1315 (\textit{22.21\%})& 194 (\textit{20.10\%})& 425 (\textit{23.76\%})  \\
Basque\textsubscript{BDT} & 5396 (\textit{33.52\%})& 1798 (\textit{33.48\%})& 1799 (\textit{31.80\%}) & 
Korean\textsubscript{Kaist} & 23010 (\textit{21.92\%})& 2066 (\textit{22.12\%})& 2287 (\textit{19.15\%})  \\
Hungarian\textsubscript{Szeged} & 910 (\textit{25.71\%})& 441 (\textit{33.56\%})& 449 (\textit{23.61\%}) &
Danish\textsubscript{DDT} & 4383 (\textit{21.83\%})& 564 (\textit{21.81\%})& 565 (\textit{18.58\%})   \\
Portuguese\textsubscript{Bosque} & 8328 (\textit{23.60\%})& 560 (\textit{19.46\%})& 477 (\textit{22.85\%}) & 
Gothic\textsubscript{PROIEL} & 3387 (\textit{16.77\%})& 985 (\textit{19.09\%})& 1029 (\textit{18.76\%})  \\
Urdu\textsubscript{UDTB}& 4043 (\textit{23.00\%})& 552 (\textit{23.01\%})& 535 (\textit{18.88\%}) & 
Lithuanian\textsubscript{HSE} & 153 (\textit{16.34\%})& 55 (\textit{16.36\%})& 55 (\textit{21.82\%})  \\ \bottomrule
\end{tabular}
\end{adjustbox}}
\caption{Total number of sentences per each dataset split (\% non-projective sentences).}
\label{tab:treebank-size}
\end{table}

\section{Label coverage}\label{label-coverage}\label{appendix-label-coverage}

At test time, our model assigns a label for each task by choosing one from a finite set learned during training. As a result, it is expected that the model may not be able to predict some of the labels occurring in the test set. Table \ref{tab:label-coverage}  reports the number of labels that have not been seen in the training and dev set and the total number of unique labels found in the test set. In addition, we include data about the percentage of occurrences of unseen labels with respect to the the occurrences of all labels in the test set. 

\begin{table}[hptb!]
\centering
\small{
\begin{adjustbox}{max width=1\linewidth}
\scalebox{0.78}{
\begin{tabular}{@{}ll
>{\columncolor[HTML]{EFEFEF}}c 
>{\columncolor[HTML]{EFEFEF}}c 
>{\columncolor[HTML]{EFEFEF}}c ccc
>{\columncolor[HTML]{EFEFEF}}c 
>{\columncolor[HTML]{EFEFEF}}c 
>{\columncolor[HTML]{EFEFEF}}c }
\toprule
 &  & \multicolumn{3}{c}{\cellcolor[HTML]{FFFFFF}\begin{tabular}[c]{@{}c@{}}Task 1\\ (1st plane)\end{tabular}} & \multicolumn{3}{c}{\cellcolor[HTML]{FFFFFF}\begin{tabular}[c]{@{}c@{}}Task 2\\ (2nd plain)\end{tabular}} & \multicolumn{3}{c}{\cellcolor[HTML]{FFFFFF}\begin{tabular}[c]{@{}c@{}}Task 3\\ (deprel)\end{tabular}} \\
\multirow{-2}{*}{Language} & \multirow{-2}{*}{Encoding} & \cellcolor[HTML]{FFFFFF}Unseen & \cellcolor[HTML]{FFFFFF}Total & \cellcolor[HTML]{FFFFFF}\begin{tabular}[c]{@{}c@{}}\% \\ occ.\end{tabular} & \cellcolor[HTML]{FFFFFF}Unseen & \cellcolor[HTML]{FFFFFF}Total & \cellcolor[HTML]{FFFFFF}\begin{tabular}[c]{@{}c@{}}\% \\ occ.\end{tabular} & \cellcolor[HTML]{FFFFFF}Unseen & \cellcolor[HTML]{FFFFFF}Total & \cellcolor[HTML]{FFFFFF}\begin{tabular}[c]{@{}c@{}}\% \\ occ.\end{tabular} \\ \midrule
 & rel-PoS &0 (\textit{0\%}) &75 & 0 & $-$ & $-$ & $-$ & 0 (\textit{0\%}) &25 & 0\\ 
 & 1p-brackets & 4 (\textit{3.39\%}) &118 & $0.02$ & $-$ & $-$ & $-$ & 0 (\textit{0\%}) &25 & 0\\
 & 2p-greedy &2 (\textit{3.08\%}) &65 & 0.01 & 1 (\textit{5.26\%}) &19 & 0.01 & 0 (\textit{0\%}) &25 & 0\\  
\multirow{-4}{*}{\begin{tabular}[c]{@{}l@{}}Ancient\\ Greek\textsubscript{Perseus}\end{tabular}} & 2p-prop &1 (\textit{1.49\%}) &67 & 0 & 0 (\textit{0\%}) &22 & 0 & 0 (\textit{0\%}) &25 & 0\\   \cmidrule(l){2-11} 
 & rel-PoS & 4 (\textit{4.3\%}) &93 &$ 0.02$ & $-$ & $-$ & $-$ & 0 (\textit{0\%}) &30 & 0\\ 
 & 1p-brackets & 2 (\textit{2.15\%}) &93 &$0.01$  & $-$ & $-$ & $-$ & 0 (\textit{0\%}) &30 & 0\\ 
 & 2p-greedy & 3 (\textit{4.35\%}) &69 & 0.02 & 2 (\textit{10.53\%}) &19 & 0.01 & 0 (\textit{0\%}) &30 & 0\\ 
\multirow{-4}{*}{Basque\textsubscript{BDT}} & 2p-prop & 3 (\textit{4.35\%}) &69 & 0.02 & 2 (\textit{10.0\%}) &20 & 0.01 & 0 (\textit{0\%}) &30 & 0\\\cmidrule(l){2-11} 
 & rel-PoS & 6 (\textit{7.5\%}) &80 & $0.06$ &$-$& $-$ & $-$ & 0 (\textit{0\%}) &47 & 0\\ 
 & 1p-brackets & 5 (\textit{6.1\%}) &82 & $0.05$ & $-$ & $-$ & $-$ & 0 (\textit{0\%}) &47 & 0\\ 
 & 2p-greedy & 1 (\textit{1.64\%}) &61 & 0.01 & 1 (\textit{8.33\%}) &12 & 0.01 & 0 (\textit{0\%}) &47 & 0\\ 
\multirow{-4}{*}{Hunagrian\textsubscript{Szeged}} & 2p-prop & 1 (\textit{1.64\%}) &61 & 0.01 & 1 (\textit{6.67\%}) &15 & 0.01 & 0 (\textit{0\%}) &47 & 0\\   \cmidrule(l){2-11} 
 & rel-PoS & 2 (\textit{2.11\%}) &95 & $0.03$ & $-$ & $-$ & $-$ & 0 (\textit{0\%}) &38 & 0\\ 
 & 1p-brackets & 0 (\textit{0\%}) &60 & 0 & $-$ & $-$ & $-$ & 0 (\textit{0\%}) &38 & 0\\ 
 & 2p-greedy & 0 (\textit{0\%}) &54 & 0 & 0 (\textit{0\%}) &14 & 0 & 0 (\textit{0\%}) &38 & 0\\ 
\multirow{-4}{*}{Portuguese\textsubscript{Bosque}} & 2p-prop &  0 (\textit{0\%}) &54 & 0 & 0 (\textit{0\%}) &14 & 0 & 0 (\textit{0\%}) &38 & 0\\ \cmidrule(l){2-11} 
 & rel-PoS &8 (\textit{7.69\%}) &104 & $0.07$ & $-$ & $-$ & $-$ & 0 (\textit{0\%}) &24 & 0\\ 
 & 1p-brackets &4 (\textit{6.06\%}) &66 & $0.03$ & $-$ & $-$ & $-$ & 0 (\textit{0\%}) &24 & 0\\ 
 & 2p-greedy & 3 (\textit{5.36\%}) &56 & 0.02 & 0 (\textit{0\%}) &12 & 0 & 0 (\textit{0\%}) &24 & 0\\ 
\multirow{-4}{*}{Urdu\textsubscript{UDTB}} & 2p-prop & 3 (\textit{5.36\%}) &56 & 0.02 & 0 (\textit{0\%}) &14 & 0 & 0 (\textit{0\%}) &24 & 0\\ \cmidrule(l){2-11} 
 & rel-PoS & 2 (\textit{2.82\%}) &71 & $0.02$ & $-$ & $-$ & $-$ & 0 (\textit{0\%}) &26 & 0\\ 
 & 1p-brackets & 4 (\textit{6.06\%}) &66 & $0.04$ & $-$ & $-$ & $-$ & 0 (\textit{0\%}) &26 & 0\\ 
 & 2p-greedy &  1 (\textit{1.92\%}) &52 & 0.01 & 1 (\textit{10.0\%}) &10 & 0.01 & 0 (\textit{0\%}) &26 & 0\\
\multirow{-4}{*}{Afrikaans\textsubscript{AfriBooms}} & 2p-prop & 1 (\textit{1.89\%}) &53 & 0.01 & 1 (\textit{9.09\%}) &11 & 0.01 & 0 (\textit{0\%}) &26 & 0\\  \cmidrule(l){2-11} 
 & rel-PoS & 1 (\textit{1.11\%}) &90 & 0 & $-$ & $-$ & $-$ & 1 (\textit{3.23\%}) &31 & 0\\ 
 & 1p-brackets & 1 (\textit{1.59\%}) &63 & 0 & $-$ & $-$ & $-$ & 1 (\textit{3.23\%}) &31 & 0\\ 
 & 2p-greedy &0 (\textit{0\%}) &56 & 0 & 0 (\textit{0\%}) &8 & 0 & 1 (\textit{3.23\%}) &31 & 0\\ 
\multirow{-4}{*}{Korean\textsubscript{Kaist}} & 2p-prop &0 (\textit{0\%}) &56 & 0 & 0 (\textit{0\%}) &8 & 0 & 1 (\textit{3.23\%}) &31 & 0\\ \cmidrule(l){2-11} 
 & rel-PoS & 2 (\textit{2.25\%}) &89 & $0.02$ & $-$ & $-$ & $-$ & 0 (\textit{0\%}) &34 & 0\\ 
 & 1p-brackets & 0 (\textit{0\%}) &72 & 0 & $-$ & $-$ & $-$ & 0 (\textit{0\%}) &34 & 0\\ 
 & 2p-greedy & 0 (\textit{0\%}) &63 & 0 & 0 (\textit{0\%}) &12 & 0 & 0 (\textit{0\%}) &34 & 0\\ 
\multirow{-4}{*}{Danish\textsubscript{DDT}} & 2p-prop & 0 (\textit{0\%}) &63 & 0 & 0 (\textit{0\%}) &12 & 0 & 0 (\textit{0\%}) &34 & 0\\  \cmidrule(l){2-11} 
 & rel-PoS & 3 (\textit{4.11\%}) &73 & $0.03$ & $-$ & $-$ & $-$ & 0 (\textit{0\%}) &31 & 0\\ 
 & 1p-brackets & 2 (\textit{2.78\%}) &72 & $0.02$ & $-$ & $-$ & $-$ & 0 (\textit{0\%}) &31 & 0\\ 
 & 2p-greedy & 1 (\textit{1.79\%}) &56 & 0.01 & 2 (\textit{13.33\%}) &15 & 0.02 & 0 (\textit{0\%}) &31 & 0\\  
\multirow{-4}{*}{Gothic\textsubscript{PROIEL}} & 2p-prop & 1 (\textit{1.79\%}) &56 & 0.01 & 2 (\textit{13.33\%}) &15 & 0.02 & 0 (\textit{0\%}) &31 & 0\\  \cmidrule(l){2-11} 
 & rel-PoS & 2 (\textit{4.17\%}) &48 & $0.19$ & $-$ & $-$ & $-$ & 1 (\textit{3.12\%}) &32 & $0.09$ \\ 
 & 1p-brackets &7 (\textit{15.91\%}) &44 & $0.66$& $-$ & $-$ & $-$ & 1 (\textit{3.12\%}) &32 & $0.09$\\ 
 & 2p-greedy &3 (\textit{8.11\%}) &37 & 0.38 & 1 (\textit{16.67\%}) &6 & 0.09 & 1 (\textit{3.12\%}) &32 & 0.09\\  
\multirow{-4}{*}{Lithuanian\textsubscript{HSE}} & 2p-prop & 3 (\textit{8.11\%}) &37 & 0.38 & 1 (\textit{16.67\%}) &6 & 0.09 & 1 (\textit{3.12\%}) &32 & 0.09\\  \hline 
\end{tabular}}
\end{adjustbox}}
\caption{Label coverage in each task at test time.} 
\label{tab:label-coverage}
\end{table}

\end{document}